\documentclass{article}
\usepackage{ijcai20}

\usepackage{times}
\usepackage{soul}
\usepackage{url}
\usepackage[hidelinks]{hyperref}
\usepackage[utf8]{inputenc}
\usepackage[small]{caption}
\usepackage{graphicx}
\usepackage{amsmath}
\usepackage{amsthm}
\usepackage{booktabs}
\usepackage{algorithm}
\usepackage{algorithmic}
\usepackage[psnames]{xcolor}
\usepackage{amssymb}
\usepackage{multirow}
\usepackage{multicol}
\definecolor{pink1}{rgb}{0.958, 0.088, 0.978}

\newcommand{\bm}[1]{\mathbf{#1}}

\title{Learning from the Scene and Borrowing from the Rich: \\Tackling the Long Tail in Scene Graph Generation}

\author{
    Tao He$^{1,2}$ \and Lianli Gao$^{2}$ \and Jingkuan Song$^2$ \and Jianfei Cai$^1$ \And Yuan-Fang Li$^1$ \footnote{Corresponding author}\\
    \affiliations
   $^1$ Faculty of Information Technology, Monash University, Australia \\
   $^2$ Center for Future Media and School of Computer Science and Engineering,
   University of Electronic Science and Technology of China, China 
    \emails
    $^1$ \{tao.he,jianfei.cai,yuanfang.li\}@monash.edu,
    $^2$ lianli.gao@uestc.edu.cn, jingkuan.song@gmail.com
}

\begin{document}

\maketitle

\begin{abstract} 
	Despite the huge progress in scene graph generation in recent years, its long-tail distribution in object relationships remains a challenging and pestering issue. Existing methods largely rely on either external knowledge or statistical bias information to alleviate this problem. In this paper, we tackle this issue from another two aspects: (1) scene-object interaction aiming at learning specific knowledge from a scene via an additive attention mechanism; and (2) long-tail knowledge transfer which tries to transfer the rich knowledge learned from the head into the tail.  
	Extensive experiments on the benchmark dataset Visual Genome on three tasks demonstrate that our method outperforms current state-of-the-art competitors. 
   
\end{abstract}

\section{Introduction}
Scene graph generation is a fundamental task in computer vision that has been successfully applied to many other tasks, including image captioning ~\cite{yang2019auto}, image retrieval~\cite{johnson2015image} and commonsense reasoning~\cite{zellers2019recognition}. 
Given an image, a relationship between objects in the image is typically denoted as a triple: $(subject, predicate, object)$, where the \emph{predicate} can also be denoted as \emph{relation}. 
To detect such relationships requires the understanding of the image content \emph{globally}. 
In scene graph generation, the representation of a relationship needs to preserve semantic information of the triple as well as the inherent attributes of the objects and the relations between them. 
It is a challenging task due to the distributional biases present in the datasets. 
For example, the benchmark dataset Visual Genome~\cite{krishna2017visual} contains $150$ distinct objects, producing possible unique relationships of approx.\ 22K. 
Such a large number of relationships are too arduous to train a model as it is impossible to cover each relationship with sufficient samples~\cite{zellers2018neural}. 
This challenge is further complicated by the highly imbalanced distribution in the relations. 
It has been observed~\cite{zellers2018neural,zhang2019large,dornadula2019visual,chen2019knowledge} that the distribution of relations of Visual Genome is highly long-tail and biased: the head relations can have $10k$ instances whereas the tail relations have less than $10$ each. 
Thus, a model can readily learn the representation of head relations but struggles to learn that of tail relations.
% However, learning such representations is a significant challenge due to the severely skewed distribution of relations. 
% \hl{\sout{Thus, the significant gap between the head and the tail makes it more challenging to generate a scene graph.
% }}
% \yf{Does the above makes sense?}

Many previous methods focus on the union region of a pair of objects~\cite{deng2014large,dai2017detecting}, where only visual features are considered, but not distributional bias of relations. 
However, as mentioned before, due to the highly imbalanced nature of relations, a relation classifier is hardly well optimized by such uneven data. 
Xu et al \shortcite{xu2017scene}  developed a message passing strategy to aid relation recognition where how to refine the object and relation feature becomes the central goal. 
However, its performance still suffers from the lack of sufficient data required for learning. 
By counting the frequency of various relations,  Neural Motifs~\cite{zellers2018neural} discovers that some relations are highly correlated with the objects. 
For instance, the possession relation ``\textsf{has}'' always exists between some specific pairs of objects, such as subject ``\textsf{man}'' and object ``\textsf{eye}''. 
Similarly based on the statistic results from a dataset, KER~\cite{chen2019knowledge} developed a knowledge routing network to preserve the relation bias into their model. 
Additionally, other work~\cite{lu2016visual} utilized  natural language information as an auxiliary tool to boost relation classification by mapping the language prior knowledge to    relation   phrases.
One limitation of these methods is their reliance on the statistic bias knowledge, without which their results would decline significantly.
Similarly, Gu et al~\shortcite{gu2019scene} leveraged ConceptNet~\cite{speer2017conceptnet}, a commonsense knowledge graph, to bridge the gap between visual features and  external  knowledge by a recurrent neural network. 

Moreover, many recent works discover that a well-represented contextual feature can significantly benefit relation recognition. 
Specifically, Graph R-CNN~\cite{yang2018graph} develops an attentional Graph Convolutional Network (aGCN), focusing on learning the contextual information between two objects that are filtered by a Relation Proposal Network (RePN).  
Qi et al~\shortcite{qi2019attentive} proposed two interacting modules to inject contextual clue to relation feature: a semantic transformer module concentrating on preserving semantic embedded relation features via projecting visual features and textual features to a common semantic space; and a graph self-attention module embedding a joint graph representation by aggregating neighboring nodes' information. Shi et al~\shortcite{shi2019explainable} utilized the attention mechanism to enhance node and relationship representation and trace the reasoning-flow in complex scene scenarios. 

In this paper we address two critical challenges in scene graph generation: (1) how to effectively encode contextual clue into its corresponding object representation; and (2) how to balance the severely skewed  predicate distribution to improve model performance.
Specifically, for the first challenge, we propose a scene-object interaction module aiming at learning the interplay coefficient between individual objects and their specific scene context. 
For instance, the relation triple ``\textsf{man} \textsf{riding} \textsf{bike}'' is usually associated with the outdoor scene instead of indoor. 
Therefore, the outdoor scene is a key contextual clue to aid us to confidently predict the ``\textsf{riding}'' relation once given the objects ``\textsf{man}'' and ``\textsf{bike}'' and the outdoor information. 
To this end, we treat annotated objects of each image as the scene label of the image and deploy a weighted multi-label classifier to learn the contextual scene clue. 
At the same time, we employ an additive attention technique to effectively fuse the clue and the objects' visual features. 
% Although LinkNet~\cite{woo2018linknet} also adopted the global image feature to refine the object representation, LinkNet is directly to concatenate them instead of focusing on the specific interaction between scene and objects.
For the second challenge, we introduce a knowledge transfer module to enhance the representation of tail (data-starved) relations, by transferring the knowledge learned in head relations to tail relations. 
In addition, we also introduce a calibration operation, inspired by the notion of reachability in reinforcement learning~\cite{savinov2018episodic}, to resize the head and tail features to enhance their features' discriminative ability. 
In summary, our contributions are threefold:

\begin{itemize}
	\item We introduce a scene-object interaction module to fuse objects' visual feature and the scene contextual clue by an additive attention mechanism.
	
	\item To alleviate the imbalanced distribution of relations, we propose a head-to-tail knowledge transfer module to preserve  rich knowledge learned from the head into the tail.
	Moreover, our calibration operation further enhances the discriminative ability of learned visual features. 
	\item We evaluate our method on the standard scene graph generation dataset Visual Genome~\cite{krishna2017visual} on three tasks: predicate classification, scene graph classification and scene graph detection, on which our model outperforms current state-of-the-art methods.
	
\end{itemize}

\section{Method}
\begin{figure*}[ht]
	\centering
	\includegraphics[width=0.9\linewidth]{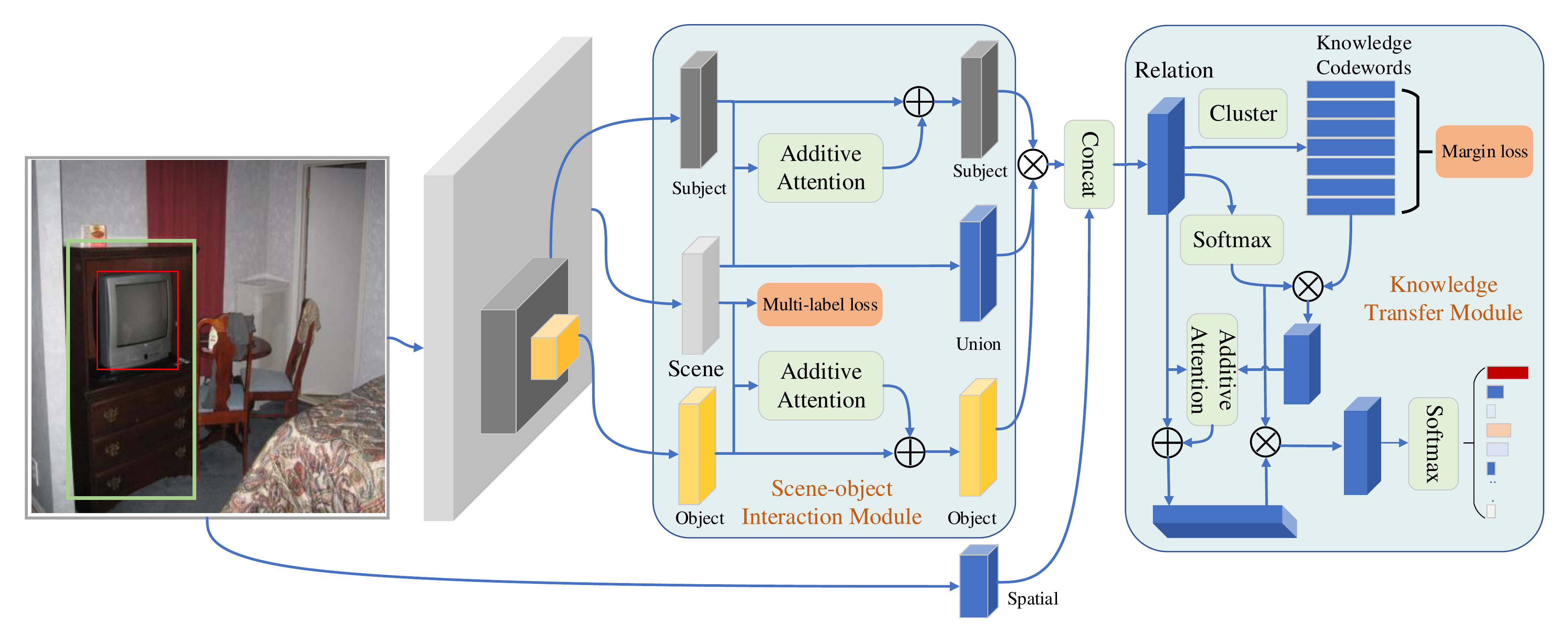}
	\caption{The high-level architecture of our framework. 
		It consists of the two main parts: the scene-object interaction module and the knowledge transfer module. 
		The scene-object interaction module refines object features by injecting the global scene interaction information. 
		The knowledge transfer module transfers the knowledge learned in the head relations to the tail relations and bridges the knowledge gap between them. 
	}
	\label{fig.frame}
\end{figure*}

Our overall framework, shown in Figure~\ref{fig.frame}, consists of three main modules: (1) feature extraction,  (2) scene-object interaction, and (3) knowledge transfer. 
Specifically, the scene-object interaction module aims to combine scene context features into object features via an additive attention mechanism, while the knowledge transfer module focuses on fusing the knowledge learned in head and tail relations to enhance their representation. 

%  Wenowlege learn will illustrate our model from four parts: feature extraction, object feature refinement via additive attention, knowledge transfer via semantic centroids and feature fusion. 
\subsection{Notations}
A scene graph is a directed relation network extracted from a multi-object image. 
Each edge in a scene graph is represented by a triple ($o_{i}$,$r_{ij}$,$o_{j}$), consisting of two objects $o_i$, $o_j$ and the relationship predicate $r_{ij}$ between them. 
%Similar to knowledge graphs~\hl{\cite{pujara2013knowledge}} \yf{Why this particular paper?} and ConceptNet~\cite{speer2017conceptnet},
Additionally, a scene graph requires to localize each object in the referring image and we denote the localization of object $o_i$ as $b_{i}$. 
Thus, given a set of object labels $\mathcal{C}$ and a set of relationship types $\mathcal{R}$ (including the none relation), a complete scene graph for an image consists of:
\begin{itemize}
	\item A set of bounding boxes $B=\{b_1, b_2, \ldots, b_n\}$, where ${b_i} \in \mathbb{R}{^4}$ denotes the coordinates of the top-left corner and the bottom-right corner, respectively.
	\item A set of objects $O=\{o_1,o_2,\ldots,o_n\}$, assigning a class label $o_i \in \mathcal{C}$ to each $b_i$. 
	\item A set of triples $T=\{(o_i,r_{ij},o_{j})\}$, where each $o_i,o_j\in O$, and $r_{ij} \in \mathcal{R}$, and that $i\neq j$. 
	%\hl{It is worth noting that we use $s_i$ to denote $o_i$ to differentiate $o_i$ and $o_j$}\yf{This is confusing}, since the order of $o_i$ and $o_j$ can render their relation different. $R$ is the set of all relation types, including the non-relation type. 
\end{itemize}

\subsection{Visual and Spatial Feature Extraction}
The first step in scene graph generation is to detect objects in an image. 
Numerous object detection methods have been proposed, e.g., Faster R-CNN~\cite{girshick2015fast}. 
To fairly compare to other baseline methods, we adopt Faster R-CNN trained on VGG-16~\cite{simonyan2014very} as our object detection and localization backbone network. 

For each detected object $o_i$, we extract two types of features: visual features $\mathbf{f}_i^o \in \mathbb{R}{^{4096}}$ and spatial features $\mathbf{l}{_i}\in \mathbb{R}{^{5}}$. 
Specifically, the visual feature extraction $\mathbf{f}_i^o$ follows that of Neural Motifs~\cite{zellers2018neural}.
The spatial features $\mathbf{l}_i$ is a 5-dimensional vector that encodes top-left and bottom-right coordinate and the size of object: $\mathbf{l}_{i}=\left[ x_{t_i}, y_{t_i}, x_{b_i}, y_{b_i}, w_i*h_i
\right]$, where $w_i$ and $h_i$ are the width and height of the object respectively. 
Recent works~\cite{zhuang2017towards,woo2018linknet} have demonstrated that the relative position of two objects in an image can significantly enhance relation recognition. 
Thus, we also encode the relative position into their relation representation as $\mathbf{s}_{ij} \in \mathbb{R}{^5}$. 
Concretely, we first convert $\mathbf{l}_i$ to the centralized coordinate as $\left[ x_{c_{i}},y_{c_i},w_i,h_i\right]$ and then calculate the relative spatial feature as 
$ \mathbf{s}_{ij}  =\left[\frac{x_{t_{j}}-x_{c_{i}}}{w_{i}}, \frac{y_{t_{j}}-y_{c_{i}}}{h_{i}}, \frac{x_{b_{j}}-x_{c_{i}}}{w_{i}}, \frac{y_{b_{j}}-y_{c_{i}}}{h_{i}}, \frac{w_{j} \cdot h_{j}}{w_{i} \cdot h_{i}}\right]
$.
It is worth noting that $\mathbf{s}_{ij}$ is different from $\mathbf{s}_{ji}$. 
To enrich the representation of $\mathbf{s}_{ij}$, we feed the above raw $5$-dimensional vector into a non-linear layer and convert it to a $256$-dimension vector $\mathbf{s}_{ij} \in \mathbb{R}^{256}$.

As for the union region features $\mathbf{f}_{ij}^u$ of subject $s_i$ and object $o_i$, we  first generate their union bounding box and follow the extraction of an object's visual feature to obtain $\mathbf{f}_{ij}^u$.

\subsection{The Scene-object Interaction Module}
% \yf{Main problem: no insights as to \emph{why} we need this interaction module}

% \het{\sout{The quality of object features is crucial for relation recognition, since the model needs not only to  recognise the relation between any pair of objects, but also to classify the objects. }}

For scene graph generation, the correct recognition of relations not only depends on object features, but also takes important cues from the scene. 
For example,  the scene of ``\textsf{outdoor}'' should contribute more to the relation ``\textsf{riding}'' while less to ``\textsf{holding}'', as riding mostly takes place in the outdoor, which is not the case for holding. 

Many works, such as  IMP~\cite{xu2017scene} and Neural Motifs~\cite{zellers2018neural}, demonstrate the contextual representation has a conspicuous effect on the relation recognition.  
% \het{\sout{we   focused on refining object features, including IMP~\cite{xu2017scene} that proposed to pass messages between different objects to iteratively refine object features, and Neural Motifs~\cite{zellers2018neural} that uses an LSTM network to encode the global context information into object features.} }
In this work, we propose a scene-object interaction module to encode the global scene contextual information into the object representation, which is implemented via an additive attention module widely used in machine translation models~\cite{bahdanau2014neural,britz2017massive}:
\begin{equation}
\mathbf{a}_{i}=\mathbf{max} \left\{0, \mathbf{w}_{g} \cdot\left(\mathbf{f}_{i}^{o}+\mathbf{f}^{s}\right)\right\}
\end{equation}
where $\mathbf{f}_i^o$ is the feature of object $o_i$, $\mathbf{f}^s$ is the global scene feature of an image, $\cdot$ denotes pointwise product,  and $\mathbf{w}_g$ computes a coefficient of interaction between the object and its contextualized scene. It is worth mentioning that all objects' features $\mathbf{f}_i^o$ in the same image share a common scene feature $\mathbf{f}^s$. $\mathbf{a}_i$ is pruned to the interval $[0, +\infty]$, and a greater value of $\mathbf{a}_i$ corresponds to more interaction with the scene, that is, the scene feature should contribute more to the object feature (see \eqref{eq:tf}). Note that LinkNet~\shortcite{woo2018linknet} has also proposed to incorporate scene features, while we consider the contribution of a scene to relations via an attention mechanism instead of a simple concatenation as in LinkNet. 
% Otherwise, $\mathbf{a}_i$ equal to zero implies the object has no interaction with the scene.
$\mathbf{w}_g$ is implemented by a fully-connected layer activated via a ReLu function. 
The global feature $\mathbf{f}^s$ is learned by a weighted multi-label classification loss: 
\begin{equation}
\label{eq.multi}
\mathcal{L}_{s}=-\sum_{c=1}^{\mathbf{|\mathcal{C}|}}\mathcal{W}_c * \mathrm{BCE}(\bm{p}_c, \bm{l}_c) 
\end{equation}
where $\mathcal{W}_c$ is a weight for each class and pre-calculated by counting the proportion of each object class in the training set, $\mathrm{p}_c$ is the probability of each class output from a sigmoid function, $\mathbf{l}_c$ is the true target label, and $\mathrm{BCE(.)}$ is a binary cross-entropy function aiming at classifying multi-label images. 
With the scene-object interaction, the object feature is then refined as:
\begin{equation}\label{eq:tf}
\tilde{\mathbf{f}}_i^o =  \mathbf{f}_i^o + \mathbf{a}_i * \mathbf{f}_i^s .
\end{equation}

From the refined object feature  $\tilde{\mathbf{f}}_i^o$, the union region feature $\mathbf{f}_{ij}^u$ and the transformed relative spatial feature $\mathbf{s}_{ij}$, we construct the final representation of each triple $(s_i,r_{ij},o_j)$ as:
\begin{equation}
\mathbf{f}_{ij}^t =\left[ \tilde{\mathbf{f}}_i^o \times \mathbf{f}_{ij}^u \times  \tilde{\mathbf{f}}_j^o; {\mathbf{s}}_{ij}  \right]
\end{equation}
where $\times$ is the element-wise multiplication following~\cite{zellers2018neural,woo2018linknet}, $\left[;\right]$ is the vector concatenation operation, and $\mathbf{f}_{ij}^t \in \mathbb{R}^{4096+256}$.

\subsection{Long-tail Knowledge Transfer}
Many previous works~\cite{zellers2018neural,chen2019knowledge} have observed that the distribution of relations is significantly unbalanced and long-tail, that very few relations (the head) have orders of magnitude more data than the majority of the relations (the tail). 
Intuitively, the head relations can be accurately classified while the less frequent relations are much more challenging. 
Therefore, how to transfer knowledge learned in the head relations to the tail is a key point in our model. 

\paragraph{Knowledge Codewords Construction.}
Inspired by the great success of knowledge transfer in domain adaptive learning~\cite{hsu2017learning,xie2018learning}, our model adopts semantic codewords as the knowledge representation for each relation class. 
Our model first learns $\mathrm{|\mathcal{R}|}$ codewords denoted as $\mathbf{D}=\{ \mathbf{d}_r \}_{r=1}^{\mathrm{|\mathcal{R}|}}$, where $\mathrm{|\mathcal{R}|}$ is the number of unique relation types. The codewords should possess two properties:  discriminative and semantic. To this end, we add two constraints to learn $\mathbf{D}$: a near-zero margin for intra-relation groups and a large margin for inter-relation groups, as follows:
\begin{equation}
\label{eq.center}
%   \mathcal{L}_d = \sum \sum_{r=1}^{|\mathbf{R}|}{( \mathbf{f}_{ij}^t - 
%   \mathbf{d}_r+ \mathbf{M}*\mathcal{Y})}\\
\mathcal{L}_d =\sum_{r=1}^{|\mathcal{R}|}  \mathcal{Y} ~\mathrm{dis}(\mathbf{f}_{ij}^t, \mathbf{d}_r) +(1- \mathcal{Y}) \max (0, M-\mathrm{dis}(\mathbf{f}_{ij}^t, \mathbf{d}_r) )
\end{equation}
where $M$ is a constant margin for inter-relation groups; $\mathcal{Y}=1$ if the relation of $\mathbf{f}_{ij}^t$ is $r$, otherwise $\mathcal{Y} = 0$; $\mathbf{d}_r$ is the learnable codewords; and $\mathrm{dis}(,)$ is a metric function to calculate two features' distance, for which we choose $L_1$ metric. 
Intuitively, $\mathcal{L}_d$ forces the same relation group to cluster together while pushes the inter-relation groups away. 

\paragraph{Knowledge Transfer.}
Relations at the tail of the distribution are hard to be trained, as there is an insufficient amount of samples for training.
Simply put, the challenge lays on the fact that feature $\mathbf{f}_{ij}^t$ learned of the tail relationships is not representative. 
Therefore, transferring knowledge learned from the head of the distribution to the tail is critical for the recognition of those data-starved relationships. 

Inspired by the hallucination strategy used in meta-learning~\cite{Zhang_2019_CVPR2,Zhang_2019_CVPR}, we propose a knowledge transfer method by hallucinating the learned features. 
Specifically, we first build a coarse classifier on $\mathbf{f}_{ij}^t$, that is,
\begin{align}
\mathbf{p} &= \mathbf{softmax}(\mathbf{f}_{ij}^t) \label{eq.prob}
% \end{equation}
% \begin{equation}
%{\mathcal{L}_p &= -\sum_{r=1}^\mathcal{|R|}\mathrm{BCE}(\mathbf{p}_r,\mathbf{l}_r)
%\label{eq:lp}}
\end{align} 
where  $\mathbf{p}$ is a probability distribution over relation types $\mathcal{R}$ implemented by a softmax classification layer. %{% with a cross-entropy loss $\mathcal{L}_p$:$\mathbf{l}_r$ is the true target label, and $\mathrm{BCE}(.)$ is the binary cross-entropy function as in Equation~\ref{eq.multi}.
Then, the hallucinated feature is calculated by:
\begin{equation}
\mathbf{\tilde{f}}_{ij}^t = { \sum_{ r=1 } ^ {\mathrm{|\mathcal{R}|}} }\mathbf{p}_{r} \mathbf{d}_{r} 
\end{equation}
where $\mathbf{d}_r$ is the informative knowledge codewords learned by Equation~\ref{eq.center}. 
Similarly, we also apply an additive attention  to combine the original feature $\mathbf{{f}}_{ij}^t$ with  the hallucinated  $\mathbf{\tilde{f}}_{ij}^t$:
\begin{equation}
\mathbf{a}_{ij}^t = \mathbf{max}\left\{0, \mathbf{w}_f \cdot (\mathbf{{f}}_{ij}^t+ \mathbf{\tilde{f}}_{ij}^t)\right\}
\end{equation}
where $\mathbf{w}_f$ is the parameters of a nonlinear layer to calculate a coefficient of two features. 
Finally, we obtain the new relation features as  the below:
\begin{equation}
\label{eq.fuse}
\mathbf{\tilde{f}}_{ij} = \mathbf{f}_{ij}^t + \mathbf{a}_{ij}^t \mathbf{\tilde{f}}_{ij}^t
\end{equation}

\paragraph{Long-tail Features Calibration.}
Ideally,  $\mathbf{\tilde{f}}_{ij}$ should be  close to $\mathbf{f}_{ij}^t$ so that the fused feature does not change $\mathbf{f}_{ij}^t$ too much, because the head relations already have sufficient samples to be trained and the codewords of head relations should be close to $\mathbf{f}_{ij}^t$. 
On the contrary, for the tail relations, the modification can be significant and arbitrary, consequentially leading to the confusion with the head relations.  

Many previous works have demonstrated that the discrimination of the head and tail class representation plays an essential role in imbalanced data learning~\cite{zhu2014capturing}.
To avoid this confusion, we calibrate $\mathbf{f}_{ij}$ to different scales for different frequency relationships by:
\begin{equation}
\mathbf{f}_{ij} = \alpha \cdot \mathbf{max(p)} \cdot \mathbf{\tilde{f}}_{ij} 
\end{equation}
where $\mathbf{p}$ is the probability vector from Equation~\ref{eq.prob}.  %and $\mathbf{max(.)}$ function is to return the maximum probability in $\mathbf{p}$. 
Generally, as for the data-rich relations, $\mathbf{max(p)}$ should be a large value, possibly close to $1$ whereas much smaller for the rare relations, because the frequent relations are trained by more data and their predicate prediction should be more confident.  
Thus, $\mathbf{max(p)}$  can be seen as a discriminative calibrating metric to separate the head and tail features. 
$\mathbf{\alpha}$ is a constant scalar to resize them. 
Finally, we deploy a relation classifier on $\mathbf{f}_{ij}$, on which a cross-entropy loss $\mathcal{L}_{rel}$ is imposed. %\yf{need to give the definition}  
\subsection{Learning}
The overall loss function is as follows:
\begin{equation}
\mathcal{L} = \mathcal{L}_s  + \mathcal{L}_{det} + \mathcal{L}_p+ \mathcal{L}_{rel} + \epsilon \mathcal{L}_d
\end{equation}
where $\mathcal{L}_s$ is a multi-label classification loss defined in Equation~\ref{eq.multi} to learn the scene feature, $\mathcal{L}_{det}$ is the object detection loss of Faster-RCNN, $\mathcal{L}_{d}$ is the knowledge codewords learning loss defined in Equation~\ref{eq.center}, $\mathcal{L}_{p}$ is the coarse relation classification loss in Equation~\ref{eq.prob}, and $\mathcal{L}_{rel}$ is the final relation classification loss defined above. 
$\mathcal{\epsilon}=0.01$ serves to balance the term of the codewords loss. 
Note that the reason why $\epsilon$ is set to a small number is that $\mathcal{L}_d$ is a distance metric usually much greater than the other terms, but not that $\mathcal{L}_d$ is not important. 
All parameters in our model are differentiable, so the model is trained in an end-to-end fashion.  

\section{Experiment}
We evaluate our method on three standard scene graph generation tasks: predicate classification (PredCls), scene graph classification (SGCls)  and scene graph detection (SGDet). 
In PredCls, given ground-truth bounding boxes and objects, the task is to predict scene graph triples on these objects. 
In SGCls, given the ground-truth bounding boxes only, the task is to predict object labels and triples.
In SGDet, the task is to localize bounding boxes, predict object labels  and triples. 

Specifically, the experiments are conducted to answer the following research questions:

\textbf{RQ1}: How does our method compare with state-of-the-art scene graph generation methods?

\textbf{RQ2}: How does each part of our model contribute to the relation recognition performance on three tasks?

\textbf{RQ3}: How well does our method perform in qualitative analysis?

\begin{table*}[!t]
	\centering
	\begin{tabular}{clcccccccccc}\toprule
		\multirow{15}{*}{\rotatebox{-90}{\small Constraint}} & \multirow{2.5}{*}{Method} & \multicolumn{3}{c}{SGDet} & \multicolumn{3}{c}{SGCls}  & \multicolumn{3}{c}{PredCls} &{\multirow{2.5}{*}{Mean}} \\  \cmidrule{3-11}
		
		& & R@20& R@50& R@100& R@20& R@50& R@100 & R@20 & R@50 & R@100& \\\midrule 
		
		& IMP& -& 3.4& 4.2& -& 21.7& 24.4& -& 44.8& 44.8& {25.3}\\ %\cmidrule{2-12} 
		
		& Graph-RCNN & -& 11.4&  13.7 & -& 21.7&  31.6& -& 54.2 &59.2 & 33.2\\ %\cmidrule{2-12} 
		
		& Neural Motifs$^\dagger$ & 20.1 & 24.8 & 27.2 & 30.2 & 33.5 & 35.5 & 52.8  &57.7  & 62.6 & 38.3\\ %\cmidrule{2-12}
		
		& Neural Motifs & 21.4 & 27.2 & 30.3 & 32.9 & 35.8 & 36.5 & 58.5 & 65.2 & 67.1 & {41.7} \\%\cmidrule{2-12} 
		
		& GSM & - & - & - & - & \textbf{38.2} & \textbf{40.4} & - & 56.6 & 61.3 & -\\ %\cmidrule{2-12} 
		
		& Mem & 7.7  & 11.4 & 13.9 & 23.3 & 27.8 & 29.5 & 42.1 & 53.2 & 57.9 & 29.6 \\%\cmidrule{2-12} 
		
		& KRE$^\dagger$ & 20.5 & 25.2 & 27.9  & 29.7 & 33.9 & 34.8 & 53.4  &58.7  &61.0 & 38.3\\ %\cmidrule{2-12} 
		
		& KRE & 22.3 & 27.1 & 29.8 & 32.3 & 36.7 & 37.4 & 59.1 &65.8 & 67.6& 42.0\\%\cmidrule{2-12}
		
		& \textbf{Ours$^\dagger$}  & 21.2 & 26.8 &  29.3 & 30.2  & 34.4  & 35.9  &  57.1 &  63.5 &  64.5 & 40.3  \\%\cmidrule{2-12}
		
		& \textbf{Ours} & \textbf{23.6} & \textbf{28.2 }& \textbf{ 31.4} & \textbf{33.6 } & {37.5 } & {38.3 } &  \textbf{60.3} &  \textbf{66.2} &\textbf{  68.0} &  \textbf{43.1 } \\ \midrule
		
		{\multirow{4}{*}{\rotatebox{-90}{\small\hspace{-6pt} Unconstraint}}} 
		& IMP & - & 22.0 & 27.4 & - & 43.4 & 47.2 & - & 75.2 & 83.6 & 49.8 \\ %\cmidrule{2-12} 
		
		& Neural Motifs&25.7 & 30.5 & 35.8 & 42.6 & 44.5 & 47.7&76.3& 81.1& 88.3& 52.5\\%\cmidrule{2-12} 
		
		& GSM & - & - & - & - & 41.4 & 46.0 & - & 61.6 & 68.9 & -\\ %\cmidrule{2-12} 
		
		& KRE& 24.6& 30.9 & 35.8 & 42.8 & 45.9 & 49.0 & 77.1& 81.9 & 88.9 & 52.9 \\ %\cmidrule{2-12} 
		
		& \textbf{Ours} & \textbf{26.9} & \textbf{31.4} & \textbf{36.5}& \textbf{43.6}& \textbf{46.2} & \textbf{50.2} & \textbf{77.9} & \textbf{82.5} & \textbf{90.2} & \textbf{53.9}\\ \bottomrule
	\end{tabular}
  \caption{Performance (R@K) comparison with the state-of-the-art methods with and without graph constraint on VG. Since some works do not test
    on R@20, we only compute the mean on the two tasks of R@50 and R@100. $\dagger$ indicates the method discards the statistical bias prior information during training.
  }
  \label{tab.sota}\end{table*}

\begin{table*}[!t]
	\centering
	\begin{tabular}{clcccccccccc}\toprule
		\multirow{9}{*}{\rotatebox{-90}{\small Constraint}} & \multirow{2.5}{*}{Method} & \multicolumn{3}{c}{SGDet} & \multicolumn{3}{c}{SGCls}  & \multicolumn{3}{c}{PredCls} &{\multirow{2}{*}{Mean}} \\  \cmidrule{3-11}
		& & R@20& R@50& R@100& R@20& R@50& R@100 & R@20 & R@50 & R@100& \\  \midrule
		& BL &  20.4 & 25.2   & 27.5 & 30.3 & 33.4   & 34.6 & 54.8 & 58.5 & 62.1 & 38.5\\ %\cmidrule{2-12} 
		& BL+SO & 22.5 & 26.7   & 30.1 & 32.5 & 35.7   & 36.8 & 58.2 & 64.2 & 66.8 & 41.5\\ %\cmidrule{2-12} 
		& BL+SO+KT &  23.0 & 27.6   & 30.9 & 33.4 & 37.1   & 38.0 & 59.8 & 65.8 & 67.6 & 42.6\\ %\cmidrule{2-12} 
		& BL+SO+KT+FC &23.6 & 28.2   & 31.4 & 33.6 & 37.5   & 38.3 & 60.3 & 66.2 & 68.0 & 43.1\\  \midrule
		
		\multirow{3.5}{*}{\rotatebox{-90}{\small\hspace{-12pt} Unconstraint}} & BL &  23.3 & 27.5   & 32.6 & 40.2 & 43.4   & 45.3 & 73.3 & 78.5 & 86.7 & 50.0\\ %\cmidrule{2-12} 
		& BL+SO & 25.4 & 29.2   & 34.3 & 42.7 & 44.7   & 48.1 & 76.4 & 80.6 & 88.0 & 52.2\\ %\cmidrule{2-12} 
		& BL+SO+KT & 26.2 & 30.7   & 35.9 & 43.1 & 45.0   & 49.4 & 77.2 & 82.1 & 89.4 & 53.3\\ %\cmidrule{2-12} 
		& BL+SO+KT+FC &  26.9 & 31.4   & 36.5 & 43.6 & 46.2   & 50.2 & 77.9 & 82.5 & 90.2 & 53.9 \\ \bottomrule
	\end{tabular}
  \caption{Ablation study results, where we study the effect of the three main modules of our method: scene-object (SO), knowledge transfer (KT) and feature calibration (FC). BL denotes the baseline without any of the above modules.}
  \label{tab.abl}
\end{table*}

\subsection{Dataset and Implementation Details}
\paragraph{Dataset.} We conduct our method on the challenging and most widely used benchmark, Visual Genome (VG)~\cite{krishna2017visual}, which consists of 108,077 images with average annotations of 38 objects and 22 relations per image. 
The experimental settings follow the previous works \cite{zellers2018neural,chen2019knowledge}, where we use   $150$ object classes for $\mathcal{C}$ and $50$ relations for $\mathcal{R}$. 
Similar to Neural Motifs~\cite{zellers2018neural}, we  utilize the statistical bias information as the extra knowledge to boost the relation recognition performance and we also report the results without this information.

\paragraph{Implementation Details.} $\alpha$ is set as $10$,  $\epsilon$ at $0.01$, and learning rate starts from $0.001$ and decays with the training processing. Codewords $\mathbf{D}=\{ \mathbf{d}_r \}_{r=1}^{\mathrm{|\mathcal{R}|}}$ is initialized by pre-calculated clusters implemented by K-means. 
We apply the Faster R-CNN ~\cite{girshick2015fast} based on VGG-16 
as the backbone object detection and localization network. 
The number of object proposals is $256$, each of which is processed by RoIAalign~\cite{he2017mask} pooling to extract object and union region features. 
We adopt the Top-K Recall (denoted as R@K) following previous work~\cite{zellers2018neural,chen2019knowledge} as the evaluation metric and report R@20, R@50 and R@100 on the three tasks.

\subsection{  Comparison with State-of-the-art Methods (RQ1) }
We compare our method to the following recent state-of-the-art methods: KRE~\cite{chen2019knowledge}, GSA~\cite{qi2019attentive}, Mem~\cite{wang2019exploring}, IMP~\cite{xu2017scene}, and Neural Motifs~\cite{zellers2018neural}. 
In addition, we also compare to Graph-RCNN~\cite{yang2018graph}, since it also develops an attention mechanism to learn contextual information. % \hl{Graph-RCNN~\cite{yang2018graph}}  \yf{Need to justify why compare with Graph-RCNN}. 
As the source code of LinkNet~\cite{woo2018linknet} is unavailable and we are unable to reproduce its results, we do not compare with LinkNet. 
It is worth noting that Neural Motifs and KRE use the relation bias as the additional prior to guide the recognition  and we report their results with or without the bias. 
Also, we report two sets of results under different conditions, constraint and unconstraint, to calculate R@K, following IMP~\cite{xu2017scene} %\yf{Need to discuss the conditions, and conditions not in this paper?}. 

Table~\ref{tab.sota} shows the results on the three tasks. 
As some methods did not report their results on the R@20, the mean result is calculated according to their reported results. 
From Table~\ref{tab.sota}, we can make the following observations. 

(1) Our method is superior to other methods in the majority of cases  even irrespective of the use of the bias information. 
Specifically, in terms of mean recall in the constraint setting, our method surpasses KRE, the best method among the baselines, by about $1.1$ percentage points when the statistical bias information is used.
A larger improvement of about $2$ percentage points is achieved when that information is not used. 
Also, the similar comparison pattern can be found in Neural Motifs. 
Compared with KRE and Neural Motifs, the performance difference between with and without statistical bias information is less in our methods ($2.8$ percentage points vs $3.7$ and $3.4$), indicating that our method does not heavily rely on this bias, and that our model can essentially learn this bias from the raw data. % instead of directly giving it to  the model. 

(2) GSM shows a great advantage in SGCls task but performs poorly in the task of the predicate classification. 
As GSM does not report the results on the scene graph detection task, we also  do not  report their mean recall. 

(3) Similarly, our method achieves the best performance in the unconstraint setting. %, and even better than GSM in the scene graph classification. 
Due to the space limitation, we do not report the result when the bias information is discarded. 
However similar observations can be made. 

\begin{table}[!t]
	\centering
	\caption{Predicate classification results of bottom-$10$ tail relations with or without the knowledge transfer module on unconstraint R@50 and R@100.}
	\label{tab.rela}
	\begin{tabular}{l c c c c }
		\toprule
		\multirow{2}{*}{Relation} & \multicolumn{2}{c}{R@50} & \multicolumn{2}{c}{R@100} \\\cmidrule(lr){2-3}\cmidrule(lr){4-5}
		& w/o KT & w KT & w/o KT & w KT \\ \cmidrule{1-5} 
		%\textsf{along}        & 5.71  &  6.32  & 8.38   &   10.42  \\ %\cmidrule{1-5}
		
		\textsf{lying on}     & 12.52 &  15.31 & 16.48  &   17.53  \\ %\cmidrule{1-5}
		\textsf{on back of}   & 3.21  &  4.86  & 6.70   &   7.58   \\ %\cmidrule{1-5}
		\textsf{to}           & 2.74  &  5.31  & 5.38   &   5.53   \\ %\cmidrule{1-5}
		\textsf{mounted on}   & 0.04  &  3.04  & 1.84   &   4.37   \\ %\cmidrule{1-5}
		\textsf{walk in}      & 3.24  &  5.53  & 5.32   &   7.28   \\ %\cmidrule{1-5}
		\textsf{across}       & 2.57  &  4.30  & 5.69   &   6.39   \\
		\textsf{made of}       & 3.56  &  3.90  & 6.69   &   6.97   \\
		\textsf{playing}       & 4.38  &  4.53  & 7.31  &   7.50   \\
		\textsf{says}       & 0.41  &  1.46  & 2.46   &   2.85 \\
		\textsf{flying in}      & 0.0 &  0.0  & 0.0   &   0.0  \\ %\cmidrule{1-5}
		\bottomrule
	\end{tabular}
\end{table}

\subsection{Effectiveness of Each Module (RQ2)}

We split our model into three modules: scene-object interaction (SO), knowledge transfer (KT) and feature calibration (FC). 
The baseline model (BL) denotes the simple model that only uses the feature generated by Faster-RCNN to recognize relations. 
The ablation study results are shown in Table~\ref{tab.abl}, where we test the performance on the three tasks by adding each module one at a time. 
For a fair comparison, all ablated models are trained by the same number of epochs, set as $40$.

We can observe that under both experimental conditions, constraint and unconstraint, the performance of the baseline is the worst. 
The addition of the scene-object interaction module SO improves the average performance by $2$--$3$ percentage points, which confirms the crucial role the global contextual information plays in relation recognition. 
When we deploy the knowledge transfer module KT, a further $1$ percentage point of improvements is gained. 
Finally, though the achievements from adding the feature calibration module FC is not as significant as the other two modules, it still obtains a noticeable lift of about $0.5$ percentage point.

Our knowledge transfer module (KT) is specifically designed to solve the problem of data imbalance. 
To evaluate its effectiveness, Table~\ref{tab.rela} shows the predicate classification (PredCls) results of bottom-$10$ tail relations whose frequencies are substantially lower than the average frequency of all relations.  
The columns ``w/o KT'' (respectively ``w KT'') denote the model without (respectively with) knowledge transfer and feature calibration. 
The superiority of the knowledge transfer module can be clearly observed. It is worth noting that since the relation \textsf{flying in} has only five samples in the entire dataset, all its results are zero.
More generally, the knowledge transfer module on average improves performance for each relation by $2$--$3$ percentage points.

Briefly, we can draw two conclusions from the ablation study. (1) The three modules all positively contribute to the relation recognition performance, and their combination achieves the best results. (2) The scene-object interaction module is the most effective of the three, as it offers more contextual clues and knowledge, and the other two modules rely on the knowledge learned from the scene context.

\begin{figure}[!t]
	\centering
	\includegraphics[width=1\linewidth]{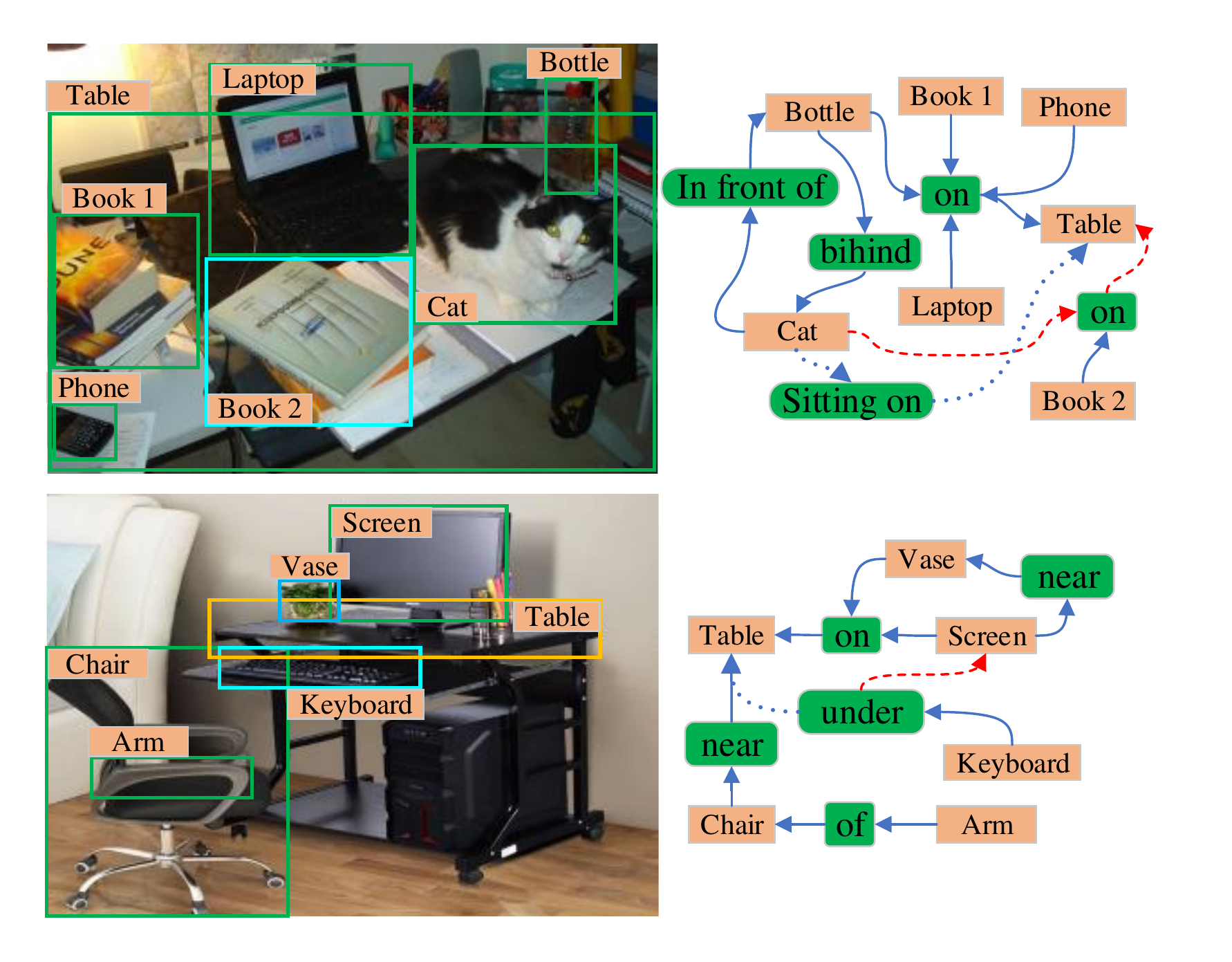}
	\caption{Qualitative results of two images based on two models: the baseline and the full model. Solid lines represent correct relations detected by both models. Dashed lines represent wrong relations detected by the baseline model. Dotted lines represent correct relations detected by the full model that the baseline model missed.}
	\label{fig.quali}
\end{figure}
\subsection{Qualitative Results (RQ3)}
Figure~\ref{fig.quali} visualizes some scene graph generation results of two models: the baseline model and the full model. 
We can observe that though the baseline model is able to capture many relations, it does get confused on some cases. 
Taking the second image as an example, the baseline model predicts that the keyboard is under the screen but in fact is under the table. 
The possible reason is that the baseline model only considers the visual and spatial feature of the screen and keyboard objects but does not consider the global scene feature. 
% Thus, it is prone to regard there is existing a relationship between screen and keyboard. 

\section{Conclusion}
In this work, we investigate the long-tail problem existing in scene graph generation. To address this issue, we propose an end-to-end framework consisting of three modules: scene-object interaction, knowledge transfer and feature calibration, each of which has its specific function. 
The extensive experimental results show that our method significantly  outperforms other state-of-the-art methods on all standard evaluation metrics. We observe that there still exists a large performance gap between the scene graph detection task and the predicate classification task. In future, we will focus on object label refinement, which is a promising way to improve scene graph generation performance.

% \section{Acknowledgement}
%  \het{This works is funded by xxx.  This works is funded by xxx. This works is funded by xxx. This works is funded by xxx. This works is funded by xxx. This works is funded by xxx. This works is funded by xxx.}

\bibliographystyle{named.bst}
\bibliography{ijcai20.bib}
\end{document}